\documentclass[10pt,twocolumn,letterpaper]{article}

\usepackage{cvpr}              

\usepackage{graphicx}
\usepackage{amsmath}
\usepackage{amssymb}
\usepackage{booktabs}

\usepackage{makecell} 
\usepackage{multirow} 

\usepackage[pagebackref,breaklinks,colorlinks]{hyperref}

\usepackage[capitalize]{cleveref}
\crefname{section}{Sec.}{Secs.}
\Crefname{section}{Section}{Sections}
\Crefname{table}{Table}{Tables}
\crefname{table}{Tab.}{Tabs.}

\def\cvprPaperID{4099} 
\def\confName{CVPR}
\def\confYear{2022}

\begin{document}

\title{Panoptic-PHNet: Towards Real-Time and High-Precision LiDAR Panoptic Segmentation via Clustering Pseudo Heatmap}

\author{Jinke Li \ \ \ \  Xiao He \  \ \  \ Yang Wen \ \ \ \   Yuan Gao \ \ \  \  Xiaoqiang Cheng \ \ \ \ Dan Zhang\\
Uisee Foundation Research \& Development\\
{\tt\small $\{$jinke.li, xiao.he, yang.wen, yuan.gao, xiaoqiang.cheng, dan.zhang$\}$@uisee.com}
}
\maketitle

\begin{abstract}
As a rising task, panoptic segmentation is faced with challenges in both semantic segmentation and instance segmentation. However, in terms of speed and accuracy, existing LiDAR methods in the field are still limited. In this paper, we propose a fast and high-performance LiDAR-based framework, referred to as Panoptic-PHNet, with three attractive aspects: 1) We introduce a clustering pseudo heatmap as a new paradigm, which, followed by a center grouping module, yields instance centers for efficient clustering without object-level learning tasks. 2) A knn-transformer module is proposed to model the interaction among foreground points for accurate offset regression. 3) For backbone design, we fuse the fine-grained voxel features and the 2D Bird's Eye View (BEV) features with different receptive fields to utilize both detailed and global information. Extensive experiments on both SemanticKITTI dataset and nuScenes dataset show that our Panoptic-PHNet surpasses state-of-the-art methods by remarkable margins with a real-time speed. We achieve the \textbf{1st place} on the public leaderboard of SemanticKITTI and leading performance on the recently released leaderboard of nuScenes.
\end{abstract}
\begin{figure}[t]
\begin{center}
\includegraphics[width=\linewidth]{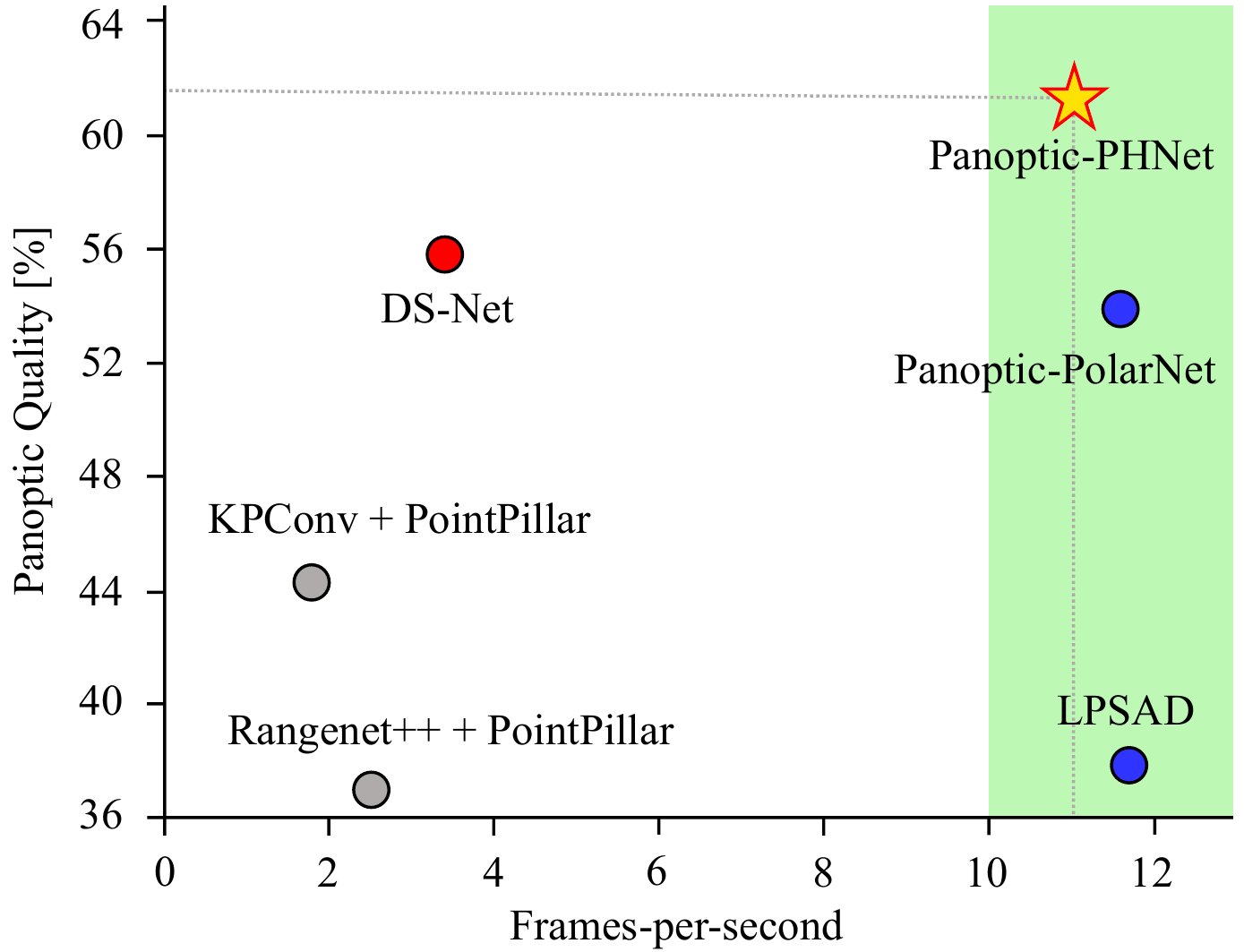}
\vspace{-2.5em}
\end{center}
   \caption{Panoptic quality vs. single frame inference latency on SemanticKITTI \cite{DBLP:behley2020benchmark}. Green area indicates real-time zone, which meets 10 frame-per-second frequency. The 2D CNN based approaches \cite{DBLP:conf/cvpr/ZhouZF21,DBLP:conf/iros/MiliotoBMS20}, the 3D CNN based approach \cite{DBLP:conf/cvpr/Hong0Z0L21} and the combined methods \cite{DBLP:behley2020benchmark} are shown in blue, red and gray respectively. Our proposed Panoptic-PHNet outperforms all other methods in PQ by a large margin and still maintains a real-time speed.}
\label{fig1}
\vspace{-1em}
\end{figure}

\section{Introduction}
\label{sec:intro}

In recent years, there has been a rapid development of autonomous driving, and as an essential perception task of its key technologies, scene understanding has attracted considerable attention from researchers. Panoptic segmentation is a recently introduced task in the image domain \cite{DBLP:conf/cvpr/KirillovHGRD19} that aims to unify semantic segmentation and instance segmentation in a single framework. With the release of LiDAR point cloud benchmarks, e.g., SemanticKITTI \cite{DBLP:behley2020benchmark} and nuScenes \cite{fong2021panoptic}, related works in the 3D field have also been extensively promoted.

The purpose of LiDAR panoptic segmentation is not only to predict class labels for all points,  including foreground points (\emph{thing}) and background points (\emph{stuff}), but also the instance IDs for \emph{thing} points. According to the implementation of instance segmentation, panoptic segmentation can be categorized into proposal-based and proposal-free approaches.

Proposal-based methods require an independent network or branch to predict proposals \cite{DBLP:behley2020benchmark,DBLP:journals/corr/abs-2004-08189}, a drawback of which is that the capability of instance segmentation heavily depends on the performance of object detection.

In contrast, proposal-free ones \cite{DBLP:conf/cvpr/Hong0Z0L21,DBLP:conf/iccv/LahoudGOP19,DBLP:conf/cvpr/PhamNHRY19} explore clustering-based methods for instance segmentation. Since such methods are not bothered with the inconsistency issue based on a cascade design, they are relatively elegant in respect of implementation. Yet the commonly used heuristic clustering algorithms, such as Mean Shift \cite{DBLP:journals/pami/ComaniciuM02} and HDBSCAN \cite{DBLP:conf/pakdd/CampelloMS13}, are time-consuming and difficult to be accelerated with GPU. Although Panoptic-PolarNet \cite{DBLP:conf/cvpr/ZhouZF21} attempts to use a lightweight instance head from Panoptic-DeepLab \cite{DBLP:conf/cvpr/ChengCZ0HAC20} to predict a center heatmap and offsets, the centers predicted by the heatmap branch may not match the clustered locations from the offset branch. The possible inconsistency between these two independent branches limits such a method.

In general,  the proposal-free methods learn the offset for each \emph{thing} point, with which the point can be shifted to be near its instance center. Projecting all the shifted points onto a BEV pseudo image, we find that it shares a similar pattern with a learned center heatmap as Panoptic-PolarNet \cite{DBLP:conf/cvpr/ZhouZF21}. In other words, the projected image can be a natural heatmap to determine the existence of instances.

Specifically, given shifted \emph{thing} points, we propose to create a \emph{clustering pseudo heatmap} by projecting these points onto a BEV image directly, the number of points in each pixel, which we call the quantitative density, represents the corresponding score. Thus, through the proposed pseudo heatmap, instance centers can be easily yielded by a window-based max-pooling and used to cluster all the \emph{thing} points as \cite{DBLP:conf/cvpr/ChengCZ0HAC20}. In this way, we remove the separate heatmap learning branch and the inconsistency issue mentioned above are hence eliminated. What you see is what you get, an object-level center is generated as long as there is a cluster of points.

Nevertheless, there are also such cases where multiple centers belonging to one instance may be generated due to inaccurate point offset regression. We further propose a \emph{center grouping module}, which integrates such redundant centers, to maintains the completeness of instances.


Moreover, it is obvious that high-quality offset regression leads to better shifted points for our clustering pseudo heatmap. Thus, inspired by Transformer \cite{DBLP:conf/nips/VaswaniSPUJGKP17}, which is popular in the natural language processing and computer vision fields, we introduce a \emph{knn-transformer module} to model the interaction of \emph{thing} points efficiently. Focusing on the relationship of spatial distance among local points,  our knn-transformer promotes the offset regression with low computational consumption. 

Regarding the design of backbone, we aggregate features at different scales more flexibly considering both accuracy and inference speed. We first extract fine-grained voxel features, which are then encoded in 2D BEV space with different receptive fields via a Unet-like network as PolarNet \cite{DBLP:conf/cvpr/0035ZDYXGF20} for real-time purpose. The 2D BEV features are further mapped back to each voxel with the height dimension. By concatenating, the obtained voxel-wise features contain not only the 2D encoded features at different BEV scales, but also the fine-grained voxel features.

We evaluate our Panoptic-PHNet on SemanticKITTI and nuScenes datasets. Extensive experiments show that our approach outperforms all the state-of-the-art methods on both of the two benchmarks (1st place on the public leaderboard of SemanticKITTI) with a real-time latency.

Our contributions are summarized as below:
\begin{itemize}
\item We propose a clustering pseudo heatmap generated from the shifted \emph{thing} points directly without extra learning tasks, which allows us to avoid the inconsistency issue between two independent branches and accelerate the clustering process. A center grouping module is further introduced to maintain the integrity of instances.
\item We propose a knn-transformer module to efficiently model the interaction among \emph{thing} points for accurate offset regression.
\item We present a backbone network fusing the fine-grained voxel features and the BEV features at different scales, which, compared with the networks utilizing pure BEV features, improves the final accuracy significantly at the cost of only a little time consumption.
\item Experiments show that our approach achieves state-of-the-art performance on both SemanticKITTI and nuScenes datasets in real time, as shown in \cref{fig1}.
\end{itemize}

\begin{figure*}[t]
\begin{center}
\includegraphics[width=\linewidth]{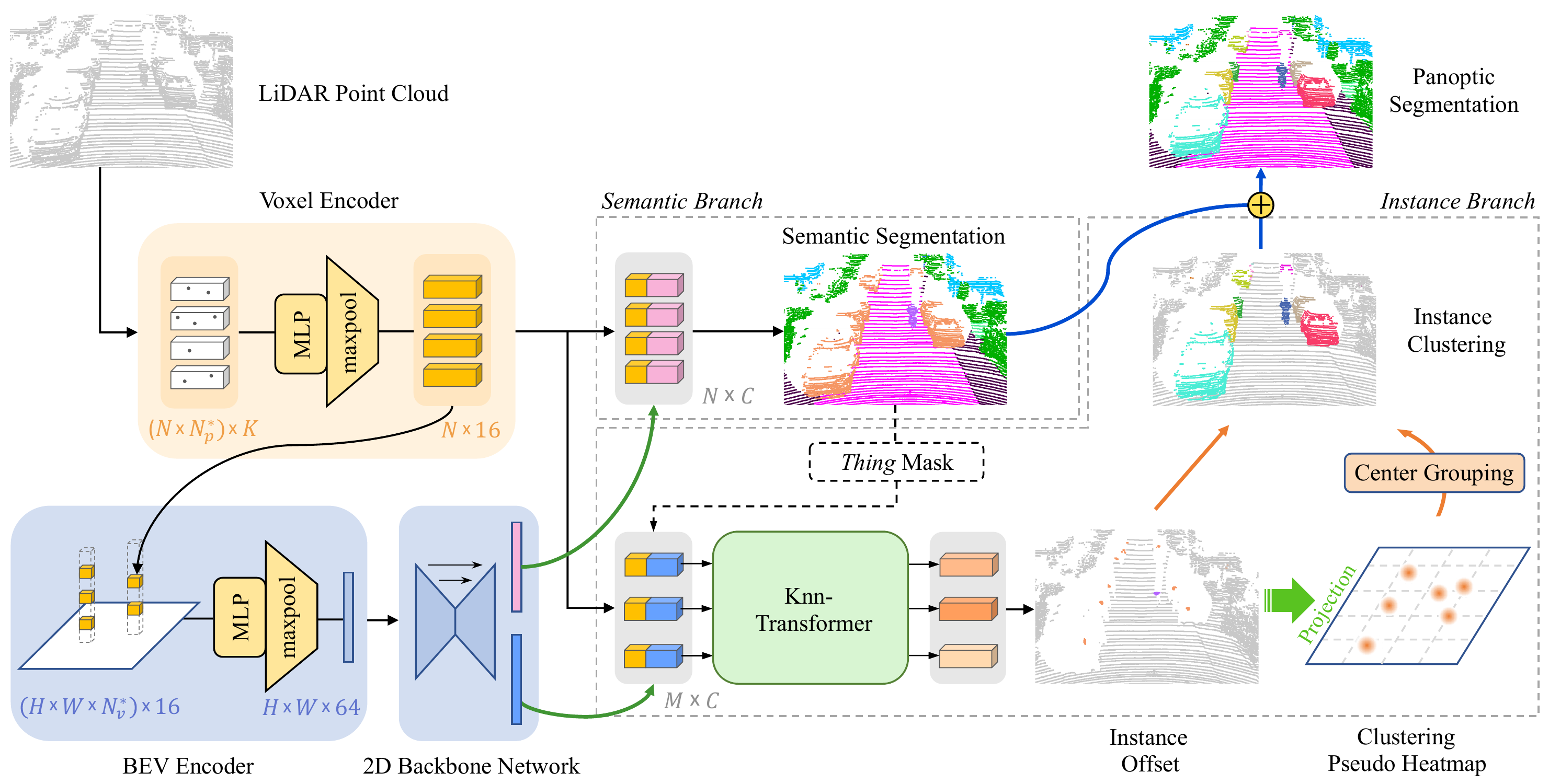}
\vspace{-2.5em}
\end{center}
   \caption{The overall framework of our Panoptic-PHNet. The backbone consists of a voxel encoder, a BEV encoder and a 2D backbone network for feature extraction. The extracted BEV features are concatenated with the fine-grained voxel features as voxel representations for semantic and instance branches. In the instance branch, a knn-transformer module is introduced to model the interaction among \emph{thing} voxels. A clustering pseudo heatmap is generated from the shifted \emph{thing} voxels to yield instance centers followed by a center grouping module. Finally, the outputs of the two branches are combined via a voting-based scheme to obtain the panoptic segmentation results.}
\label{fig2}
\vspace{-0.5em}
\end{figure*}

\section{Related Work}
\label{sec:rela}

\subsection{Representation Learning of Point Cloud}
As effective data representation is the foundation of learning based tasks, for irregular and sparse point clouds, there are generally two ways to learn representations in previous studies. One is to learn features directly at the point level, while the other regularizes raw point clouds at first before feature extraction. Based on PointNet \cite{DBLP:conf/cvpr/QiSMG17} and PointNet++ \cite{DBLP:conf/nips/QiYSG17}, KPConv \cite{DBLP:conf/iccv/ThomasQDMGG19} and RandLA \cite{DBLP:conf/cvpr/Hu0XRGWTM20} process the irregular point clouds straightforwardly, which, however, takes a time-consuming preprocess to build the graph among points. VoxelNet \cite{DBLP:conf/cvpr/ZhouT18} first projects point clouds to regular voxels and utilizes 3D CNNs to learn features. SECOND \cite{DBLP:journals/sensors/YanML18} introduces sparse convolution to promote learning efficiency for voxel-wise features. To further optimize the latency of feature extraction as well as the memory consumption, PointPillars \cite{DBLP:conf/cvpr/LangVCZYB19} collapses the height dimension via PointNet and then treats the input as a BEV image. PolarNet \cite{DBLP:conf/cvpr/0035ZDYXGF20} takes the imbalanced distribution of points in physical space into consideration and encodes point clouds into a polar BEV map. Range image \cite{DBLP:journals/corr/abs-2103-10039,DBLP:conf/icra/WuZZYK19,DBLP:conf/iros/MiliotoVBS19,DBLP:conf/icra/WuWYK18} is another common projection space for efficient feature encoding, yet the 3D topological relations are also weakened. Methods \cite{DBLP:conf/eccv/WangFKRPFS20,DBLP:conf/corl/ZhouSZAGOGNV19,DBLP:conf/eccv/TangLZLLWH20} propose to fuse information from different perspectives.

\subsection{LiDAR Panoptic Segmentation}
As a newly proposed research domain, panoptic segmentation unifies semantic segmentation and instance segmentation. In terms of  the approaches of processing ID information, two frameworks, i.e., proposal-based and proposal-free, are designed.

\noindent \textbf{Proposal-based panoptic segmentation.} PanopticTrackNet \cite{DBLP:journals/corr/abs-2004-08189} utilizes Mask R-CNN \cite{DBLP:conf/iccv/HeGDG17} for instance segmentation, and attaches a semantic head to classify the \emph{stuff} points. SemanticKITTI \cite{DBLP:behley2020benchmark} and nuScenes \cite{fong2021panoptic} release LiDAR panoptic segmentation datasets and report results by jointing existing state-of-the-art object detectors and semantic segmentation networks. For proposal-based methods, although the predicted bounding boxes make it easy to segment instances, the final performance depends heavily on the object detection task.

\noindent \textbf{Proposal-free panoptic segmentation.}
Panoptic-PolarNet \cite{DBLP:conf/cvpr/ZhouZF21} adopts a lightweight instance head from Panoptic-DeepLab \cite{DBLP:conf/cvpr/ChengCZ0HAC20} to predict instance centers and point offsets without bounding box regression. However, it is still limited due to the inconsistency issue between two independent branches and the possible failures of instance center prediction. There are also studies \cite{DBLP:conf/cvpr/Hong0Z0L21, DBLP:journals/ral/GasperiniMMNT21,DBLP:conf/iros/MiliotoBMS20} using pure clustering for instance segmentation. DS-Net \cite{DBLP:conf/cvpr/Hong0Z0L21} proposes a dynamic shifting module to shift points towards the instance centers in an iterative manner and utilizes the Mean Shift clustering to segment instances. It should be noted that taken as post process, conventional clustering methods are usually time-consuming.


\section{Methodology}
\label{sec:metho}

\subsection{Overview}

As pixels are the essential elements for Unet, voxels are the basic units in our network. Following the design of Unet, which fuses low-level and high-level features for each pixel, our network aggregates both 2D semantic features under different receptive fields and fine-grained 3D features for each voxel. The former quickens the convergence of the task and the latter facilitates distinguishing different voxels from each other.

The framework of our Panoptic-PHNet is shown in \cref{fig2}. The input LiDAR point cloud is first encoded through a voxel encoder to be 3D voxel representations, which are further transformed into size-fixed 2D representations by a BEV encoder. A Unet-like 2D backbone network is utilized to extract BEV features with different receptive fields. The BEV features are gathered for each voxel according to their coordinates. New features are generated by concatenating the gathered BEV features with the low-level fine-grained voxel features, which are then fed into two branches for semantic and instance segmentation respectively. In the instance branch, a knn-transformer module is introduced for modeling the interaction among \emph{thing} voxels to enhance the feature representation. We predict offsets towards instance centers to shift \emph{thing} voxels, subsequently a clustering pseudo heatmap is generated by projection according to the quantitative density of shifted voxels to yield instance centers. The possible redundant centers are integrated through a center grouping module.
At last, combining the outputs of two branches, we obtain the final panoptic segmentation results via a voting-based scheme as \cite{DBLP:conf/cvpr/ZhouZF21}. In the following sections, we first elaborate on two components in the instance branch of our Panoptic-PHNet, then the backbone design is presented.

\subsection{Clustering Pseudo Heatmap}
\label{sec:cph}

After the center offsets prediction and the voxels shifting, the remaining work in the instance branch can be regarded as a clustering task. For efficient instance clustering, we proceed in the BEV space based on the assumption as \cite{DBLP:conf/cvpr/ZhouZF21} that the interested \emph{thing} objects are separated from each other and do not overlap under the bird's eye view.

To address the current issues of existing methods as analyzed in \cref{sec:intro}, we propose a clustering pseudo heatmap to yield instance centers by projecting the shifted \emph{thing} voxels onto a BEV map without extra learning branches. Since the number of voxels is used as the score for each BEV grid, the location with the most voxels in a local area corresponds to a local peak on the pseudo heatmap, which can be naturally taken as an instance center. Such a bottom-up design ensures the consistency between the instance centers and the shifted voxels. More specifically, we project the shifted \emph{thing} voxels $V^{\prime} \in \mathbb{R}^{M \times 3}$ to a BEV pseudo image $I^{\prime} \in \mathbb{R}^{H \times W \times C_n}$ based on their locations on the BEV map, where $M$ is the number of \emph{thing} voxels, $H$ and $W$ are the size of the BEV map and $C_n$ is the number of the semantic categories. By summing the voxel number along the dimension of $C_n$, a class-agnostic pseudo heatmap $I \in \mathbb{R}^{H \times W \times 1}$ is created. As a non-maximum suppression, a window-based 2D max pooling is adopted to efficiently pick out the local centers. Compared with the dense learning-based heatmap in \cite{DBLP:conf/cvpr/ZhouZF21}, our clustering pseudo heatmap is sparse so that the top-$k$ operation for centers filtering is no longer necessary. Finally, each shifted \emph{thing} voxel can be clustered to its closest center according to their spatial distance on the BEV space. We use the grid size 0.2m $\times$ 0.2m for our clustering pseudo heatmap in all the experiments.

\begin{figure}[t]
\begin{center}
\includegraphics[width=\linewidth]{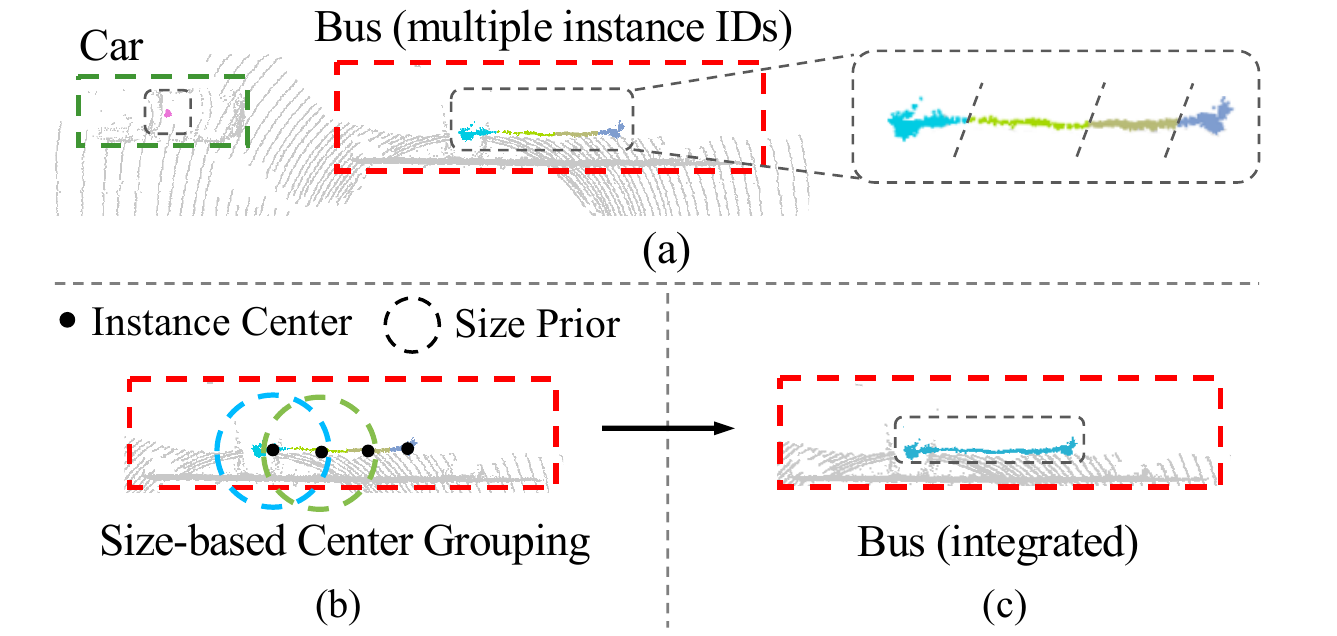}
\vspace{-2.8em}
\end{center}
   \caption{(a) illustrates a bad offset regression for a bus that is close to the origin of the LiDAR coordinate. The shifted \emph{thing} points in different color represent different instance IDs. (b) and (c) show that with our center grouping module, the bus can be appropriately integrated.}
\label{fig3}
\vspace{-1.0em}
\end{figure}

\noindent \textbf{Center grouping.} By further analyzing the results of the offset regression, it is observed that clustering itself works worse for big objects such as buses than small objects such as cars and persons. The reason is that by LiDAR sensors, usually less body part can be scanned for a big object, especially when it is close to the LiDAR origin. As illustrated in \cref{fig3} (a), the \emph{thing} points of the bus are clustered into four instance IDs instead of one as expected. In other words, multiple centers originally belonging to the same instance may be generated from the pseudo heatmap.

To deal with such cases, we introduce a size-based center grouping module. We first use a 2D average pooling on the pseudo image $I^{\prime} \in \mathbb{R}^{H \times W \times C_n}$ to count the number of \emph{thing} voxels within a sliding window for each category. Majority voting is applied to determine the class of each grid. We then give each center a minimum radius empirically based on its category. \cref{fig3} (b) illustrates the grouping operation: given a certain base center $C_b \in \mathcal{ID}_b$ as well as a radius $r_b$, where $\mathcal{ID}_b$ indicates a group of centers with the same instance ID as $C_b$. If a target center $C_t \in \mathcal{ID}_t$ with the same semantic label as $C_b$ appears within the radius $r_b$, the set $\mathcal{ID}_t$ are then regrouped into $\mathcal{ID}_b$. We traverse all centers with this operation, through which multiple redundant centers are integrated together as shown in \cref{fig3} (c). Since the number of  instance centers per LiDAR frame is limited, the time cost can almost be ignored.

\subsection{Knn-Transformer}

At the beginning of the instance branch, we use the \emph{thing} mask generated from semantic segmentation to pick out the feature vectors $F \in \mathbb{R}^{M \times C}$ for \emph{thing} voxels. Since the number of the voxels to be processed is significantly reduced, accurately modeling the interaction among these elements becomes possible. Similar to a natural sentence, the disorderd and number-unfixed \emph{thing} voxels are suitable for Transformer \cite{DBLP:conf/nips/VaswaniSPUJGKP17} to deal with.

\begin{figure}[t]
\begin{center}
\includegraphics[width=\linewidth]{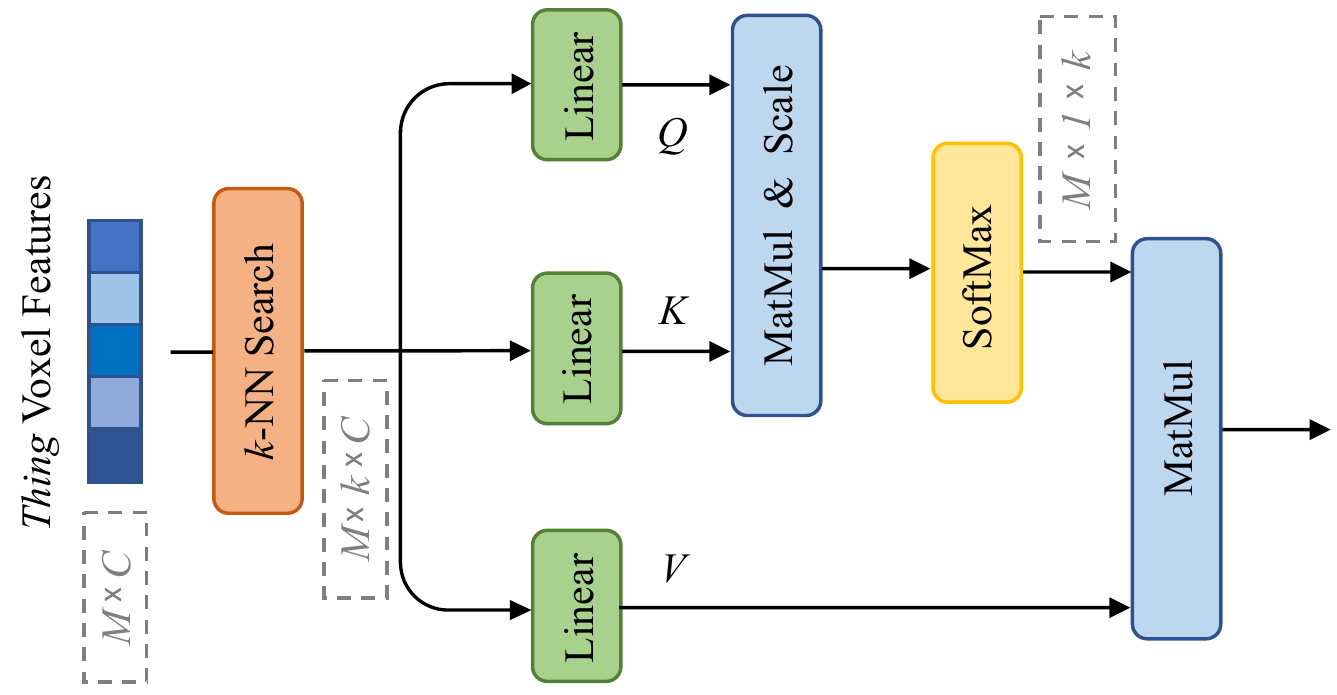}
\vspace{-1em}
\end{center}
   \caption{The self-attention layer in our knn-transformer module.}
\label{fig4}
\vspace{-1em}
\end{figure}

Inspired by Swin-Transformer \cite{liu2021Swin} where the local attention mechanism is introduced, we propose a knn-transformer to efficiently model the interaction among \emph{thing} voxels. We basically follow the design of self-attention layer from \cite{DBLP:conf/nips/VaswaniSPUJGKP17}, except that we take spatial distance as prior to construct the similarity matrix among local \emph{thing} voxels. More specifically, given the features of \emph{thing} voxels with shape $M \times C$ as shown in \cref{fig4}, we calculate the indices of $k$ nearest neighbors for each \emph{thing} voxel on GPU based on its spatial location, by which the input vectors are broadcasted to be a feature matrix with shape $M \times k \times C$. Through linear transformation, a query matrix $Q$, a key matirx $K$ and a value matix $V$ are generated respectively. Afterwards, an attention matrix with shape $M \times k$, describing the interaction between each \emph{thing} voxel and its $k$ nearest neighbors, is computed as \cite{DBLP:conf/nips/VaswaniSPUJGKP17}:
\vspace{-0.3em}
\begin{equation}
\vspace{-0.2em}
\operatorname{Attention}(Q, K, V)=\operatorname{softmax}\left(\frac{Q K^{T}}{\sqrt{C^{\prime}_{}}}\right) V
\end{equation}
where $C^{\prime}$ denotes the size of the channel dimension. Compared with the vanilla self-attention layer, our knn-based design reduces the computational complexity per layer from $O\left(M^{2} \cdot C^{\prime}\right)$ to $O\left(M \cdot k \cdot C^{\prime}\right)$. We maintain the structure of multi-head attention and feed-forward layer in \cite{DBLP:conf/nips/VaswaniSPUJGKP17}. Since the position information is encoded in each voxel by nature, the positional embedding is not adopted in our module. We use $k = 25$ in our experiments.

\subsection{Backbone Design}

\noindent \textbf{Space partition.} In our 2D backbone network, we utilize the polar BEV coordinate for space partition based on two reasons. First, objects not only remain unchanged in scale but also rarely overlap \cite{DBLP:conf/cvpr/ZhouZF21} in the BEV space. Second, the distribution of points under different ranges can be balanced in the polar coordinate \cite{DBLP:conf/cvpr/0035ZDYXGF20}. To facilitate the projection process from 3D to 2D space, cylindrical space partition \cite{DBLP:conf/cvpr/Zhu0WHM00L21} is adopted for voxel feature extraction.

\noindent \textbf{Voxel encoder.} Following \cite{DBLP:conf/cvpr/Zhu0WHM00L21}, we first group a frame of raw LiDAR point cloud into a voxel representation with shape $\left(N \times N_{p}^{*}\right) \times K$ based on the location of each point in the cylindrical space, where $K$ is the feature dimension of LiDAR points, $N$ is the number of non-empty voxels and $N_{p}^{*}$ denotes the different number of points in each voxel. A shared three-layer MLP with BatchNorm and ReLU is utilized to extract point features, followed by a max-pooling layer to create a consistent representation for each voxel. A single-layer MLP is adopted for feature reduction to generate the fine-grained voxel features with shape $N \times 16$.

\noindent \textbf{BEV encoder and 2D backbone network.} 
We futher encode features with different receptive fields in the 2D BEV space. On the one hand, the operations of interaction in 3D space are time-consuming and cost large memories. On the other hand, for a 2.5D scene scanned by LiDAR \cite{DBLP:conf/cvpr/HuZHR20}, it is unnecessary to extract features entirely in 3D space. Precisely, we first map the fine-grained voxel features $V \in \mathbb{R}^{N \times 16}$ to a polar BEV image $I_{p}^{\prime} \in \mathbb{R}^{\left(H \times W \times N_{v}^{*}\right) \times 16}$, where $H$ and $W$ are the size of the BEV map and $ N_{v}^{*}$ is the different number of voxels in each BEV grid. A shared MLP is applied for feature extraction. Similar with the voxel encoder, we use a max-pooling layer at each BEV grid to create a consistent representation $I_{p} \in \mathbb{R}^{H \times W \times 64}$. Subsequently, a Unet-like 2D backbone network is adopted to encode features with different receptive fields in the BEV space as \cite{DBLP:conf/cvpr/ZhouZF21}.

We have four decoders in our 2D backbone network, where the first two are shared for the semantic and instance branches. We gather the BEV features for the corresponding voxels in both branches respectively according to their BEV locations. The gathered BEV features are then concatenated with the fine-grained voxel features as the final voxel representation. All prediction results and supervision signals are at the voxel level. Finally, we map the voxel results into point level based on the coordinate of each point.

\begin{table*}[t]\footnotesize
\begin{center}
\setlength{\tabcolsep}{2.5mm}{
\begin{tabular}{l|cccc|ccc|ccc|c|c}
\Xhline{0.8pt}
Method & PQ & PQ$^{\dagger}$ & RQ & SQ & PQ$^{\text{Th}}$ & RQ$^{\text{Th}}$ & SQ$^{\text{Th}}$ & PQ$^{\text{St}}$ & RQ$^{\text{St}}$ & SQ$^{\text{St}}$ & mIoU & FPS \\
\hline
RangeNet++ \cite{DBLP:conf/iros/MiliotoVBS19} + PointPillars \cite{DBLP:conf/cvpr/LangVCZYB19} & 37.1 & 45.9 & 47.0 & 75.9 & 20.2 & 25.2 & 75.2 & 49.3 & 62.8 & 76.5 & 52.4 & 2.4 \\
LPSAD \cite{DBLP:conf/iros/MiliotoBMS20} & 38.0 & 47.0 & 48.2 & 76.5 & 25.6 & 31.8 & 76.8 & 47.1 & 60.1 & 76.2 & 50.9 & 11.8 \\
KPConv \cite{DBLP:conf/iccv/ThomasQDMGG19} + PointPillars \cite{DBLP:conf/cvpr/LangVCZYB19} & 44.5 & 52.5 & 54.4 & 80.0 & 32.7 & 38.7 & 81.5 & 53.1 & 65.9 & 79.0 & 58.8 & 1.9 \\
Panoster \cite{DBLP:journals/ral/GasperiniMMNT21} & 52.7 & 59.9 & 64.1 & 80.7 & 49.4 & 58.5 & 83.3 & 55.1 & 68.2 & 78.8 & 59.9 & - \\
Panoptic-PolarNet \cite{DBLP:conf/cvpr/ZhouZF21} & 54.1 & 60.7 & 65.0 & 81.4 & 53.3 & 60.6 & 87.2 & 54.8 & 68.1 & 77.2 & 59.5 & 11.6 \\
DS-Net \cite{DBLP:conf/cvpr/Hong0Z0L21} & 55.9 & 62.5 & 66.7 & 82.3 & 55.1 & 62.8 & 87.2 & 56.5 & 69.5 & 78.7 & 61.6 & 3.2${\dagger}$ \\
EfficientLPS \cite{DBLP:journals/corr/abs-2102-08009} & 57.4 & 63.2 & 68.7 & 83.0 & 53.1 & 60.5 & 87.8 & \textbf{60.5} & \textbf{74.6} & 79.5 & 61.4 & - \\
\hline
Panoptic-PHNet & \textbf{61.5} & \textbf{67.9} & \textbf{72.1} & \textbf{84.8} & \textbf{63.8} & \textbf{70.4} & \textbf{90.7} & 59.9 & 73.3 & \textbf{80.5} & \textbf{66.0} & 11.0 \\
Panoptic-PHNet $\S$ & 64.6 & 70.2 & 74.9 & 85.7 & 66.9 & 73.3 & 91.5 & 63.0 & 76.1 & 81.5 & 68.4 & - \\
\Xhline{0.8pt}
\end{tabular}
}
\end{center}
\vspace{-2.0em}
\caption{LiDAR panoptic segmentation resuts on the \textbf{test} set of SemanticKITTI. Metrics are in [$\%$] and FPS is in [Hz]. (${\dagger}$: we measure the latency of \cite{DBLP:conf/cvpr/Hong0Z0L21} with official codebase released by the authors on our hardware for reference; $\S$: our method with double-flip and multi-model ensemble.)}
\label{tab1}
\vspace{-0.8em}
\end{table*}

\begin{table*}[t]\small
\begin{center}
\setlength{\tabcolsep}{2.4mm}{
\begin{tabular}{l|cccc|ccc|ccc|c}
\Xhline{0.8pt}
Method & PQ & PQ$^{\dagger}$ & RQ & SQ & PQ$^{\text{Th}}$ & RQ$^{\text{Th}}$ & SQ$^{\text{Th}}$ & PQ$^{\text{St}}$ & RQ$^{\text{St}}$ & SQ$^{\text{St}}$ & mIoU \\
\hline
EfficientLPS \cite{DBLP:journals/corr/abs-2102-08009} & 62.4 & 66.0 & 74.1 & 83.7 & 57.2 & 68.2 & 83.6 & 71.1 & 84.0 & 83.8 & 66.7 \\
Panoptic-PolarNet \cite{DBLP:conf/cvpr/ZhouZF21} & 63.6 & 67.1 & 75.1 & 84.3 & 59.0 & 69.8 & 84.3 & 71.3 & 83.9 & 84.2 & 67.0 \\
SPVNAS \cite{DBLP:conf/eccv/TangLZLLWH20} + CenterPoint \cite{DBLP:conf/cvpr/YinZK21} & 72.2 & 76.0 & 81.2 & 88.5 & 71.7 & 79.4 & 89.7 & 73.2 & 84.2 & 86.4 & 76.9 \\
Cylinder3D++ \cite{DBLP:conf/cvpr/Zhu0WHM00L21} + CenterPoint \cite{DBLP:conf/cvpr/YinZK21} & 76.5 & 79.4 & 85.0 & 89.6 & 76.8 & 84.0 & 91.1 & 76.0 & \textbf{86.6} & 87.2 & 77.3 \\
(AF)$^{2}$-S3Net \cite{DBLP:conf/cvpr/ChengRTLL21} + CenterPoint \cite{DBLP:conf/cvpr/YinZK21} & 76.8 & 80.6 & 85.4 & 89.5 & 79.8 & 86.8 & 91.8 & 71.8 & 83.0 & 85.7 & 78.8 \\
\hline
Panoptic-PHNet & \textbf{80.1} & \textbf{82.8} & \textbf{87.6} & \textbf{91.1} & \textbf{82.1} & \textbf{88.1} & \textbf{93.0} & \textbf{76.6} & \textbf{86.6} & \textbf{87.9} & \textbf{80.2} \\
Panoptic-PHNet $\S$ & 81.5 & 84.0 & 88.4 & 91.9 & 83.5 & 88.7 & 93.9 & 78.2 & 87.8 & 88.6 & 81.5 \\
\Xhline{0.8pt}
\end{tabular}
}
\end{center}
\vspace{-2.0em}
\caption{LiDAR panoptic segmentation resuts on the \textbf{test} set of nuScenes. All scores are in [$\%$]. ($\S$: our method with double-flip and multi-model ensemble.)}
\label{tab2}
\vspace{-1.0em}
\end{table*}

\section{Experiments}
\label{sec:exper}

We evaluate our proposed Panoptic-PHNet on both SemanticKITTI and nuScenes datasets. Due to page limitations, please refer to the supplementary material for more details on the experiments and qualitative results.

\noindent \textbf{SemanticKITTI.} SemanticKITTI \cite{DBLP:behley2020benchmark} is the first dataset that presents challenges for LiDAR panoptic segmentation. It is derived from KITTI \cite{DBLP:conf/cvpr/GeigerLU12} odometry dataset, and contains 22 data sequences with a 64-beams LiDAR sensor, 10 of which are for training, 11 for testing and 1 for validation. There are annotated point-wise labels in 20 classes for segmentation tasks, 8 of which are defined as \emph{thing} classes.

\noindent \textbf{NuScenes.} NuScenes \cite{DBLP:conf/cvpr/CaesarBLVLXKPBB20} is a large-scale driving dataset with a wide diversity of urban scenes. It contains 1000 scenes of 20s duration. The annotations are created every 0.5s with a 32-beams LiDAR sensor. Recently, the official expanded the point-wise annotations for LiDAR panoptic segmentation task with 16 semantic classes, 10 of which are \emph{thing} classes. Since no one has reported results on the nuScenes test server for this new task yet, we mainly compare our results with strong baselines reported by the official \cite{fong2021panoptic} on the test and validation sets.

\noindent \textbf{Evaluation Metrics} As defined in \cite{DBLP:conf/cvpr/KirillovHGRD19}, we use panoptic quality (PQ), segmentation quality (SQ) and recognition quality (RQ) to evaluate panoptic segmentation. These metrics are calculated separately for \emph{thing} and \emph{stuff} classes indicated by $\text{PQ}^{\text{Th}}$, $\text{SQ}^{\text{Th}}$, $\text{RQ}^{\text{Th}}$ and $\text{PQ}^{\text{St}}$, $\text{SQ}^{\text{St}}$, $\text{RQ}^{\text{St}}$. Following \cite{DBLP:conf/cvpr/PorziBCK19}, we also report $\text{PQ}^{\dagger}$ to use SQ as PQ for \emph{stuff} classes. We use mean IoU (mIoU) to evaluate the quality of semantic segmentation. In addition, we adopt average EPE (end-point-error) from the visual optical flow field as the metric for offset regression to compare our approach with other clustering-based methods.

\noindent \textbf{Training and Inference.} We use the same configuration and training schedules as previous works \cite{DBLP:conf/cvpr/ZhouZF21, DBLP:conf/cvpr/Zhu0WHM00L21}. See supplementary material for detailed hyper-parameters. During training, we use the  cross-entropy loss ($\mathcal{L}_{ce}$) and Lovasz softmax loss \cite{DBLP:conf/cvpr/BermanTB18} ($\mathcal{L}_{ls}$) to train the semantic head. In the instance head, we use L1 loss ($\mathcal{L}_{l1}$) for the offset regression. The final loss is denoted as:
\begin{equation}
\mathcal{L}=\mathcal{L}_{ce}+\mathcal{L}_{ls}+\mathcal{L}_{l1}
\end{equation}
As mentioned in \cite{DBLP:conf/cvpr/ZhouZF21}, we also find that the number of dynamic instances in SemanticKITTI is limited, hence we adopt a copy-paste data augmentation scheme from \cite{DBLP:journals/sensors/YanML18} to alleviate the distribution imbalance among categories. On nuScenes, however, we do not use this data augmentation scheme, for the reason that each frame in nuScenes has 34 instances on average, which is 6 times more than that of SemanticKITTI and enough to drive the training of semantic head. During inference, we follow \cite{DBLP:conf/cvpr/ZhouZF21} to merge the results from two branches to generate the final panoptic segmentation predictions. The inference latency is measured on a platform with an Intel Core i7 CPU and a RTX 2080Ti GPU.


\begin{table*}[t]\small
\begin{center}
\setlength{\tabcolsep}{3.1mm}{
\begin{tabular}{l|cccc|ccc|ccc|c}
\Xhline{0.8pt}
Method & PQ & PQ$^{\dagger}$ & RQ & SQ &PQ$^{\text{Th}}$ & RQ$^{\text{Th}}$ & SQ$^{\text{Th}}$ & PQ$^{\text{St}}$ & RQ$^{\text{St}}$ & SQ$^{\text{St}}$ & mIoU \\
\hline
PanopticTrackNet \cite{DBLP:journals/corr/abs-2004-08189} & 51.4 & 56.2 & 63.3 & 80.2 & 45.8 & 55.9 & 81.4 & 60.4 & 75.5 & 78.3 & 58.0 \\
EfficientLPS \cite{DBLP:journals/corr/abs-2102-08009} & 62.0 & 65.6 & 73.9 & 83.4 & 56.8 & 68.0 & 83.2 & 70.6 & 83.6 & 83.8 & 65.6 \\
Panoptic-PolarNet \cite{DBLP:conf/cvpr/ZhouZF21} & 63.4 & 67.2 & 75.3 & 83.9 & 59.2 & 70.3 & 84.1 & 70.4 & 83.5 & 83.6 & 66.9 \\
\hline
Panoptic-PHNet & \textbf{74.7} & \textbf{77.7} & \textbf{84.2} & \textbf{88.2} & \textbf{74.0} & \textbf{82.5} & \textbf{89.0} & \textbf{75.9} & \textbf{86.9} & \textbf{86.8} & \textbf{79.7} \\
\Xhline{0.8pt}
\end{tabular}
}
\end{center}
\vspace{-2.0em}
\caption{LiDAR panoptic segmentation resuts on nuScenes validation. All scores are in [$\%$].}
\label{tab3}
\end{table*}


\begin{figure*}[t]
\begin{center}
\includegraphics[width=\linewidth]{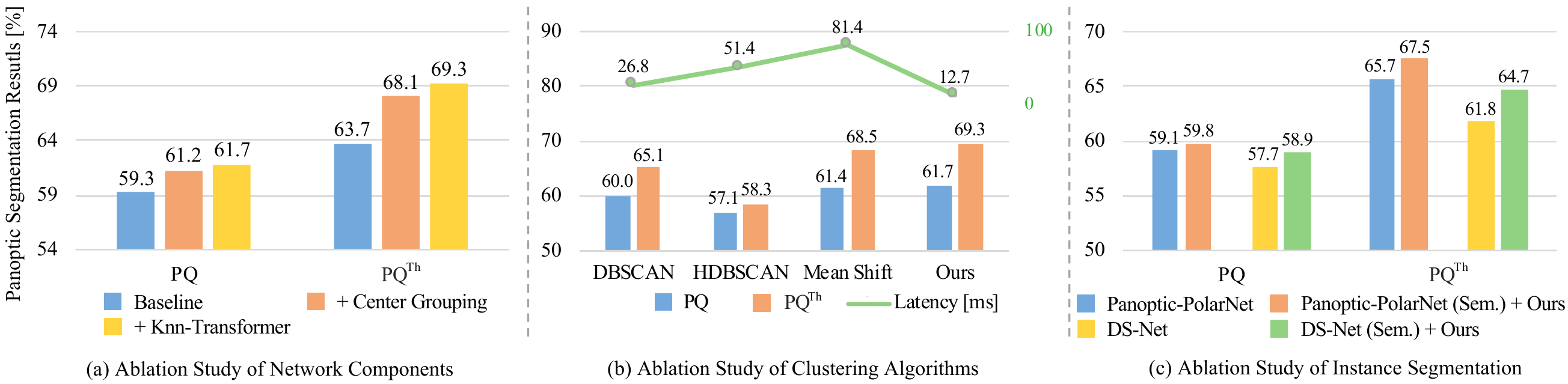}
\vspace{-2em}
\end{center}
   \caption{Ablation study on SemanticKITTI validation. (a) The network benefits from the two proposed components. (b) Our approach based on the clustering pseudo heatmap is faster and more accurate. (c) Fed with the same results of semantic segmentation respectively, our instance segmentation performs better than the two state-of-the-art LiDAR panoptic segmentation methods.}
\label{fig5}
\vspace{-0.3em}
\end{figure*}

\subsection{Main Results}

\noindent \textbf{Results on SemanticKITTI.} We first compare our method with the state-of-the-art LiDAR panoptic segmentation methods on SemanticKITTI test set. As shown in \cref{tab1}, our method outperforms all existing methods with remarkable margins, i.e., improving PQ by 4.1$\%$  (61.5$\%$ vs. 57.4$\%$) and PQ$^{\text{Th}}$ by 8.7$\%$ (63.8$\%$ vs. 55.1$\%$) with a real-time speed. Compared with 3D CNN based methods, such as DS-Net \cite{DBLP:conf/cvpr/Hong0Z0L21}, our approach achieves higher accuracy and is more than 3 times faster (11 FPS vs. 3.2 FPS). In regard of the 2D CNN based methods, e.g., Panoptic-PolarNet \cite{DBLP:conf/cvpr/ZhouZF21}, our method achieves over 10$\%$ promotion in PQ$^{\text{Th}}$ with a similar inference speed due to the combination of fine-grained voxel features and 2D CNN features. Moreover, following \cite{DBLP:conf/cvpr/YinZK21}, we also report our test-time-augmentation (TTA) version including double-flip and multi-model ensemble for reference (last line of \cref{tab1}) to show the upper bound of our framework.

\noindent \textbf{Results on NuScenes.}
Recently nuScenes released the test server for LiDAR panoptic segmentation along with the results of multiple strong baselines \cite{fong2021panoptic}. As shown in \cref{tab2}, our method surpasses the best baseline method by 3.3$\%$ PQ, 2.3$\%$ PQ$^{\text{Th}}$ and 4.8$\%$ PQ$^{\text{St}}$. Since the official combined approaches are obtained by downloading individual submissions from various evaluation servers, which may use incorporated TTA such as Cylinder3D++ \cite{DBLP:conf/cvpr/Zhu0WHM00L21}, we also report our TTA version as we do on SemanticKITTI. Although the combined methods based on the powerful CenterPoint in \cref{tab2} perform relatively better on nuScenes, we argue that such top-down approaches, depending heavily on detectors, are limited as analyzed in \cref{sec:intro}. Based on elaborative architecture design, our end-to-end framework still achieves leading results.

Following the recent official evaluation protocol \cite{fong2021panoptic}, we further report our results on nuScenes validation as shown in \cref{tab3} without any test-time augmentation techniques. Our approach outperforms the best baseline Panoptic-PolarNet \cite{DBLP:conf/cvpr/ZhouZF21} by 11.3$\%$ PQ and 14.8$\%$ PQ$^{\text{Th}}$.

\subsection{Ablation Study}

\noindent \textbf{Ablation on Network Components.} We first analyze the proposed center grouping module for our clustering pseudo heatmap as well as the knn-transformer, which are enabled sequentially. The baseline result comes from our Panoptic-PHNet model without these two modules.

As shown in \cref{fig5} (a), both of the two modules contribute to the final performance. Regarding the center grouping module, it presents a significant quality improvement for instance segmentation (+4.4$\%$ PQ$^{\text{Th}}$). As mentioned in \cref{sec:cph}, our clustering pseudo heatmap indeed achieves high recall for center generation: as long as there is a cluster of points, a highlight peak shows up certainly. However, it also brings the issue of multiple redundant centers. Solving this problem effectively, the proposed center grouping further ensures the high precision for instance centers. Note that a possible limitation of this module lies in the introduced hyper-parameters, i.e., the prior size for each category, which can be left in further work.

In addition, by enhancing the feature representations of \emph{thing} voxels for more accurate offset regression, our knn-transformer further improves the performance by 1.2$\%$ PQ$^{\text{Th}}$, further analysis is shown later.

\noindent \textbf{Ablation on Clustering Algorithms.} We compare our clustering scheme, i.e., the center grouping based clustering pseudo heatmap, with the widely used heuristic clustering algorithms: DBSCAN \cite{DBLP:conf/kdd/EsterKSX96}, HDBSCAN \cite{DBLP:conf/pakdd/CampelloMS13} and Mean Shift \cite{DBLP:journals/pami/ComaniciuM02} in a plug-in manner as \cite{DBLP:conf/cvpr/Hong0Z0L21}. \cref{fig5} (b) shows our clustering scheme outperforms all the listed clustering algorithms in both accuracy and speed. In contrast with the Mean Shift that shows the best accuracy among the heuristic algorithms, ours is more than 6 times faster (12.7ms vs. 81.4ms). It is noteworthy that in terms of accuracy, our method is just 0.8$\%$ PQ$^{\text{Th}}$ higher than Mean Shift, which is quite different from the experiments of DS-Net \cite{DBLP:conf/cvpr/Hong0Z0L21}: their dynamic shifting clustering algorithm surpasses Mean Shift by 3.2$\%$ PQ$^{\text{Th}}$. The reason inside lies in the different bases of offset regression, our high-quality offset prediction itself leads to remarkable performance even with Mean Shift, which we explain in detail later. 

\noindent \textbf{Ablation on Instance Segmentation.} We further figure out the performance of our instance segmentation compared with two clustering-based methods, i.e., DS-Net which is based on a dynamic shifting module to shift \emph{thing} points towards the instance centers in an iterative manner,  as well as Panoptic-PolarNet \cite{DBLP:conf/cvpr/ZhouZF21} requiring a heatmap by learning to yield instance centers. To eliminate the influence of the semantic segmentation, we replace the results of our semantic branch with theirs. In this way, our framework generates instance IDs for their \emph{thing} points. As shown in \cref{fig5} (c), our instance segmentation brings improvements of 1.8$\%$ and 2.9$\%$ in PQ$^{\text{Th}}$ for the two state-of-the-art methods.

\noindent \textbf{Ablation on Fine-grained Voxel Features.} We separately verify the influence of the fine-grained voxel features on the semantic branch and the instance branch. In ablation for the semantic branch, we follow Panoptic-PolarNet \cite{DBLP:conf/cvpr/ZhouZF21} to use BEV features only to generate multiple predictions, which are reshaped back to voxels. As shown in \cref{tabFeature}, the use of fine-grained voxel features improves mIoU by 1.2$\%$, and the improved mIoU further promotes PQ by 1.1$\%$ and PQ$^{\text{Th}}$ by 1.7$\%$. It can be observed that the fine-grained voxel features are crucial for the semantic segmentation task. As for the instance branch, the fine-grained voxel features are also beneficial although not so obvious as it is in the semantic branch. Overall, despite a little computational load, the combination of fine-grained voxel features and BEV features with different receptive fields brings our strong backbone network in terms of accuracy and speed for LiDAR panoptic segmentation.

\begin{table}[t]\small
\begin{center}
\setlength{\tabcolsep}{3.0mm}
\begin{tabular}{llccc}
\Xhline{0.8pt}
Branch & Method &PQ  & PQ$^{\text{Th}}$ & mIoU \\
\hline
\multirow{2}{*}{Semantic} & BEV Feature & 60.6 & 67.6 & 64.5 \\
& + Voxel Feature & \textbf{61.7} & \textbf{69.3} & \textbf{65.7} \\
\hline
\multirow{2}{*}{Instance} & BEV Feature & 61.6 & 69.0 & - \\
& + Voxel Feature & \textbf{61.7} & \textbf{69.3} & - \\
\Xhline{0.8pt}
\end{tabular}
\begin{center}
\vspace{-1.5em}
\caption{Ablation on fine-grained voxel features. We show the experiments for the semantic branch and the instance branch respectively on SemanticKITTI validation.}
\label{tabFeature}
\end{center}
\vspace{-3.0em}
\end{center}
\end{table}

\begin{table}[t]\small
\begin{center}
\setlength{\tabcolsep}{2.4mm}
\begin{tabular}{lcccc}
\Xhline{0.8pt}
Method     & PQ  & PQ$^{\text{Th}}$ & mIoU &EPE $\downarrow$ \\
\hline
DS-Net \cite{DBLP:conf/cvpr/Hong0Z0L21} & 57.7 & 61.8 & 63.5 &64.3 \\
Panoptic-PolarNet \cite{DBLP:conf/cvpr/ZhouZF21} & 59.1 & 65.7 & 64.5 & 44.9 \\
\hline
Ours  & \textbf{61.7} & \textbf{69.3} & \textbf{65.7} & \textbf{13.7} \\
\Xhline{0.8pt}
\end{tabular}
\begin{center}
\vspace{-1.5em}
\caption{Comparison among clustering-based methods on SemanticKITTI validation. EPE is calculated by shifted \emph{thing} points for each method. Panoptic metrics are in [$\%$] and EPE is in [cm]. $\downarrow$ is for lower better.}
\label{tabepe}
\end{center}
\vspace{-3.0em}
\end{center}
\end{table}

\begin{figure}[t]
\begin{center}
\includegraphics[width=\linewidth]{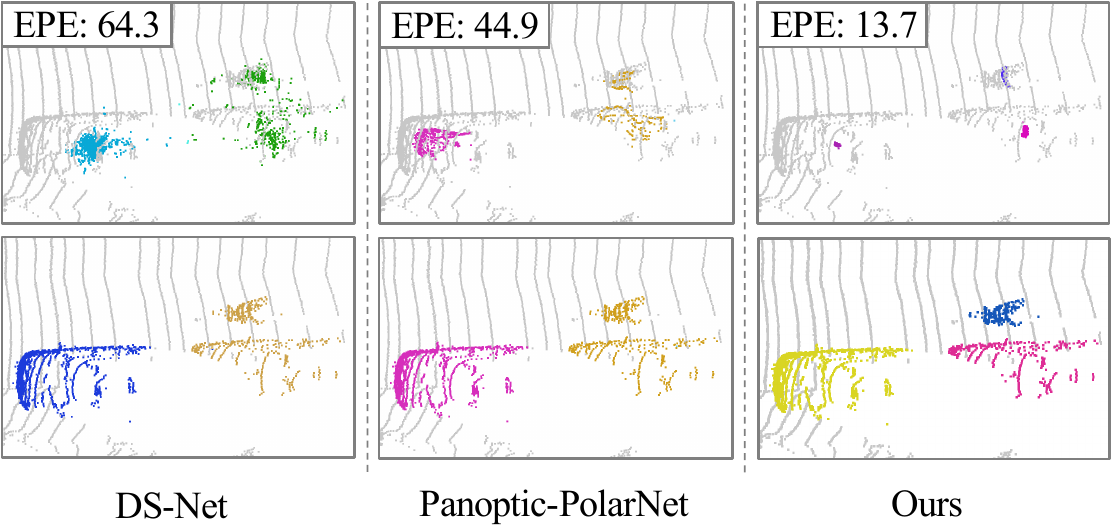}
\end{center}
\vspace{-1.5em}
   \caption{Contrast of qualitative results. The top three images show the shifted \emph{thing} points via the predicted offsets from different methods. The bottom images show the results of instance segmentation. Colored points represent different instance IDs. In this case, the three close instances are correctly segmented only with our method. Better viewed in color and zoomed in for details.}
\vspace{-0.5em}
\label{fig7}
\end{figure}

\subsection{Further Analysis}

\noindent \textbf{Effects of Offset Regression.} We further analyze the accuracy of offset regression. It is actually a key factor in clustering-based panoptic segmentation methods, for which, however, related evaluation researches can rarely be found. Thus, we adopt the average EPE (end-point-error), which is used in the evaluation of the visual optical flow field, to validate the effects of offset regression.

We still compare our method with DS-Net \cite{DBLP:conf/cvpr/Hong0Z0L21} and Panoptic-PolarNet \cite{DBLP:conf/cvpr/ZhouZF21}. Since the latter only has 2D offset predictions, all the EPE results in \cref{tabepe} are calculated in the cartesian 2D BEV space for a fair comparison. As shown in the table, our approach has an obviously smaller offset error, which means larger spatial distance among instances to facilitate the following clustering process. Although the index of mIoU also contributes to PQ, it is clear that our network benefits from such high-precision offset predictions. Moreover, high-quality offset regression presents greater impact on crowded urban scenes as illustrated in \cref{fig7}, where only our approach segments three close instances correctly. Such crowded scenes are more common in nuScenes dataset, on which, thus, our method shows more significant improvement. As demonstrated in \cref{tab3} and \cref{tabepe}, for example, compared with Panoptic-PolarNet, the promotions of our approach are 14.8\% PQ$^{\text{Th}}$ vs. 3.6\% PQ$^{\text{Th}}$ on the two datasets respectively.

\begin{figure}[t]
\begin{center}
\includegraphics[width=\linewidth]{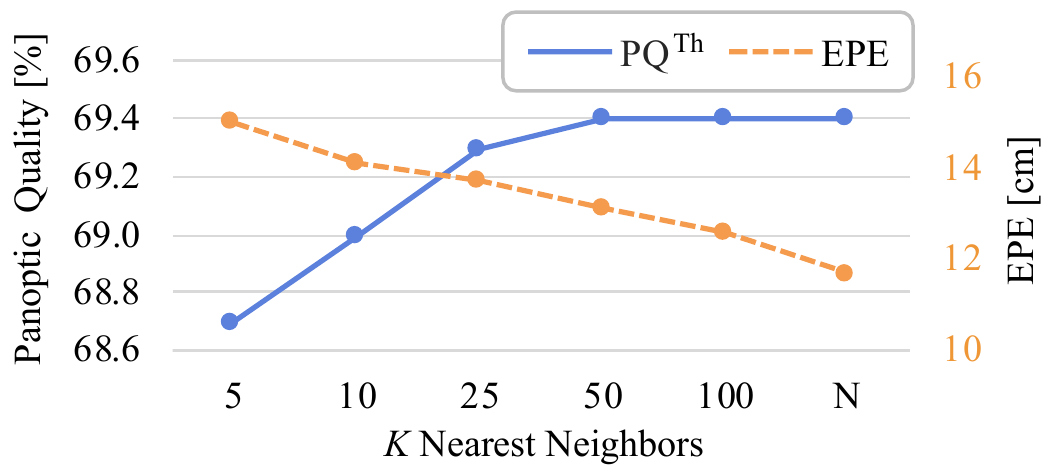}
\end{center}
\vspace{-1.5em}
   \caption{Effects of the selected \emph{k} values in knn-transformer. The primary y-axis on the left is for PQ$^{\text{Th}}$, and the secondary y-axis on the right is for EPE.}
\vspace{-1.0em}
\label{fig6}
\end{figure}

\noindent \textbf{Effects of Knn-Transformer.} We evaluate the effects of a set of \emph{k} values for our knn-transformer. As shown in \cref{fig6}, along with the increase of \emph{k}, the PQ$^{\text{Th}}$ improves at first, but almost saturates when \emph{k} is greater than 25. The EPE of offset regression, however, decreases continuously, which proves the effectiveness of our knn-transformer module. Considering the memory footprint, we use $k = 25$ as the final choice.


\section{Conclusion}

In this paper, we propose a real-time and high-precision LiDAR panoptic segmentation framework named Panoptic-PHNet. As a new paradigm, we present a clustering pseudo heatmap, which is directly generated from shifted \emph{thing} points without extra learning tasks to yield instance centers, followed by a center grouping module towards the issue of multiple redundant centers. A knn-transformer is introduced to efficiently model the interaction among \emph{thing} voxels for feature enhancement to improve the accuracy of offset regression. Finally, based on a strong backbone design, which fuses the fine-grained voxel features and 2D BEV features with different receptive fields, our Panoptic-PHNet achieves the state-of-the-art performance on both SemanticKITTI and nuScenes datasets.

{\small
\bibliographystyle{ieee_fullname}
\bibliography{egbib}
}

\ 

\ 

\ 

\ 

\ 

\ 

\ 

\ 

\ 

\ 

\ 

\ 

\ 

\ 

\ 

\ 

\ 

\ 

\ 

\ 

\ 

\ 

\ 

\ 

\ 

\ 

\ 

\ 

\ 

\ 

\ 

\ 

\ 

\ 

\ 

\ 

\newpage
\appendix

\section{Implementation Details}

In nuScenes, annotations are created every 0.5s, we follow the common practice \cite{DBLP:conf/cvpr/YinZK21,DBLP:journals/sensors/YanML18,DBLP:conf/cvpr/CaesarBLVLXKPBB20} to transform the Lidar points of non-annotated frames into their following annotated frames to generate denser point clouds, which improves our model by 2.3\% PQ on nuScenes validation. For data augmentation, we use global scaling with a random factor within [0.95, 1.05], global rotation with a random factor within [$-\pi$/2, $\pi$/2] and random flipping along both X and Y axes of the LiDAR coordinate on both datasets. As mentioned in our paper, we also use copy-paste data augmentation scheme from \cite{DBLP:journals/sensors/YanML18} on SemanticKITTI to alleviate the distribution imbalance among categories.

Our model is trained with a total batch size of 16 on 8 RTX3090 GPUs. To save computation, we train the model for 40 epochs following \cite{DBLP:conf/cvpr/Zhu0WHM00L21} for semantic segmentation and then train the instance branch for another 20 epochs. All the submissions and ablation experiments are conducted with the same setting.

For our test-time-augmentation version, we follow \cite{DBLP:conf/cvpr/YinZK21} to apply flip testing, which improves PQ by 0.3$\%$ on nuScenes validation. We ensemble five models with inputs of 3D cylindrical size from [240, 180, 32] to [576, 448, 32], which further improves PQ by 1.2$\%$. Note that we only use our single model without any TTA techniques for all the comparisons in our paper, the TTA version is just for reference considering some of methods incorporate TTA. Our single-model version achieves the 1st place on the public leaderboard of SemanticKITTI.

Regarding the hyper-parameters of the center grouping module, we calculate the average size for different categories with corresponding 3D bounding box annotations on KITTI \cite{DBLP:conf/cvpr/GeigerLU12} and nuScenes respectively. For a certain \emph{thing} category with the average size [width, length, height], we assign it a radius $r = \text{min}(\text{width}, \text{length})$.

For supervision signals, voxel-wise losses are adopted for both semantic and instance branches. We obtain voxel-wise semantic labels by majority-voting and use the mean offsets of points as the voxel-wise offset labels. We use centers of axis-aligned bounding boxes as instance centers to train the offsets, which is explained later.

\begin{figure}[t]
\begin{center}
\includegraphics[width=\linewidth]{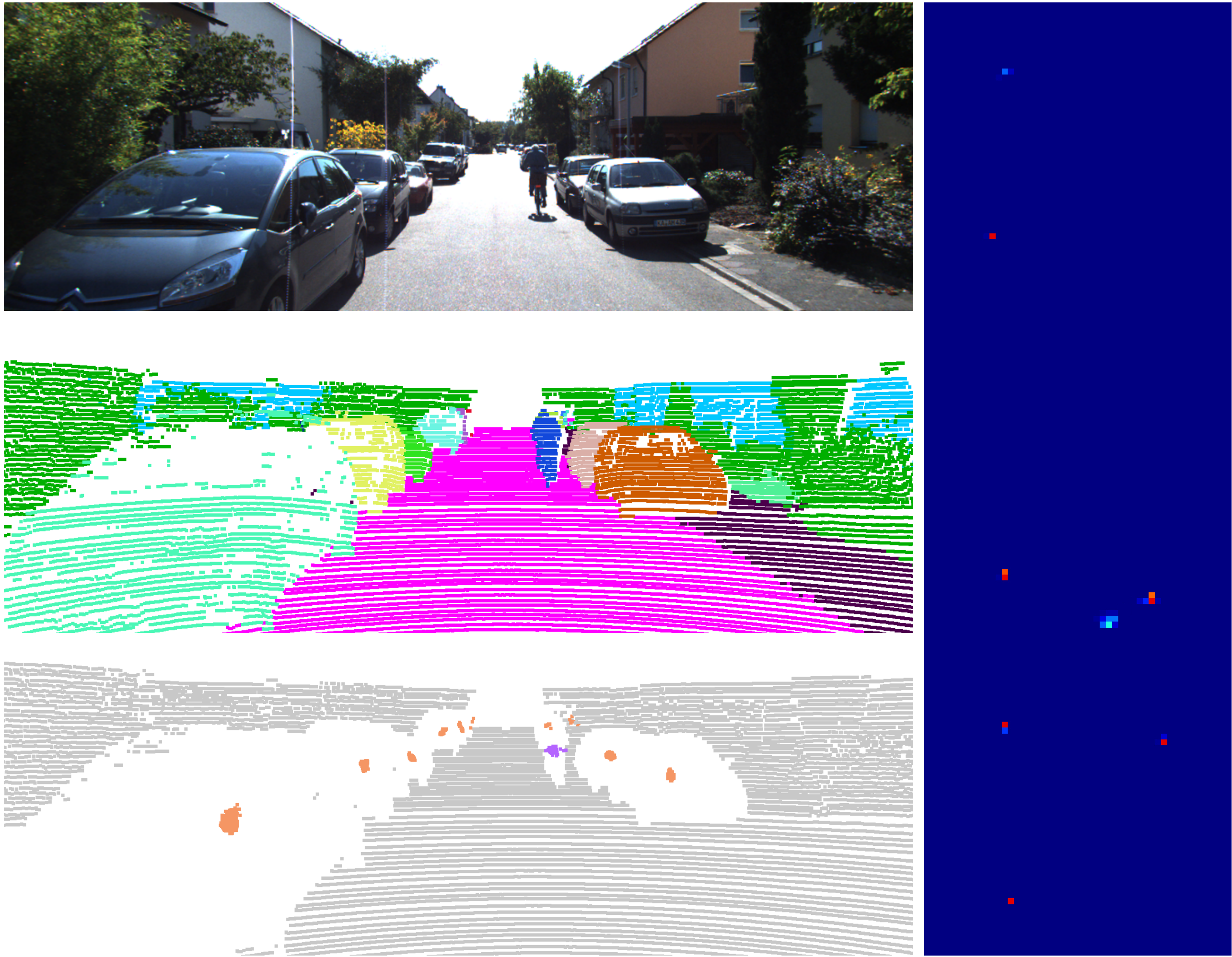}
\vspace{-2.5em}
\end{center}
   \caption{Qualitative example on SemanticKITTI. The left three images are the front view image, panoptic segmentation result and shifted \emph{thing} points. The right image is the corresponding BEV clustering pseudo heatmap.}
\label{fig_sub1}
\vspace{-1em}
\end{figure}

\section{Discussion}

\textbf{Pseudo Heatmap vs. Learned Heatmap.} We compare our clustering pseudo heatmap (PHM) with the learned heatmap (LHM) adopted in Panoptic-PolarNet \cite{DBLP:conf/cvpr/ZhouZF21}. We follow \cite{DBLP:conf/cvpr/ZhouZF21} to train a heatmap head (with their post-processing) in our framework. \cref{labelPHM} shows our PHM outperforms LHM on both datasets, especially on nuScenes (+6.8$\%$ PQ$^{\text{Th}}$). For LHM, there may be inconsistencies in terms of quantity and location between the predicted centers and the clusters of shifted \emph{thing} points. In crowded scenes (commonly seen in nuScenes, while few in semanticKITTI validation), such inconsistency issue has more impacts. Note that the mIoU drops from 77.5 to 67.4 after being refined by LHM centers on nuScenes. It's a strong evidence that LHM is quite inaccurate. On the contrary, our PHM is created from the projection of the shifted \emph{thing} points, where highlights show up certainly as long as there are clustered ones, as illustrated in \cref{fig_sub1}, so that object-level high recall is achieved. It should be noted that the more accurate the offset regression is, the sparser the clustering pseudo heatmap becomes. As a result, our PHM performs much better on nuScenes.

\begin{table}[h]\footnotesize
\begin{center}
\begin{tabular}{llllll}
\Xhline{0.8pt}
Dataset & Method &PQ  & PQ$^{\text{Th}}$ & mIoU & mIoU*\\
\hline
\multirow{2}{*}{sem.KITTI} & LHM & 61.1 & 67.9 & 65.2 & 65.1 \\
& PHM (ours) & \textbf{61.7} & \textbf{69.3 (+1.4)} & 65.2 & 65.1 \\
\hline
\multirow{2}{*}{nuScenes} & LHM & 69.1 & 65.7 & \textcolor[RGB]{255, 0, 0}{67.4} & \textcolor[RGB]{255, 0, 0}{77.5} \\
& PHM (ours) & \textbf{73.4} & \textbf{72.5 (+6.8)} & 77.5 & 77.5 \\
\Xhline{0.8pt}
\end{tabular}

\caption{Pseodo heatmap vs. learned heatmap. (mIoU*: original semantic results. mIoU: the semantic results refined by instance IDs)}
\label{labelPHM}

\begin{center}
\end{center}
\vspace{-5.5em}
\end{center}
\end{table}

One more thing, a possible weakness of PHM is that there may be multiple center predictions for one object as the shifted \emph{thing} points are not concentrated enough. Fortunately, our proposed center grouping module provides a effective solution.

\textbf{Choice of Instance Center.} There are two types of instance center used in previous researches as the supervision signals of offset regression, i.e., the mass center \cite{DBLP:conf/cvpr/ZhouZF21} and the axis-aligned center \cite{DBLP:conf/cvpr/Hong0Z0L21}. In addition, since there are 3D bounding box annotations in nuScenes as external data, the centers of bounding boxes can also be taken as instance centers. We conduct contrast experiments on both SemanticKITTI and nuScnes datasets for these three choices. As shown in \cref{tabcenter}, the difference between the mass center and the axis-aligned center is not obvious on SemanticKITTI. On nuScenes, however, the axis-align center outperforms the mass center by 2.1$\%$ PQ, and the annotated center only further improves PQ by 0.1$\%$. It is clear that the annotated center is most beneficial to offset regression due to the highest consistency. Since we do not use external data on nuScenes for comparisons, we adopt axis-aligned center as the final choice. The different results on the two datasets lie in the fact that there are plenty of crowded scenes with more dynamic object instances in nuScenes, where the choice of higher consistent centers performs better.

\begin{table}[h]\small
\begin{center}
\setlength{\tabcolsep}{1.8mm}
\begin{tabular}{lcccc}
\Xhline{0.8pt}
Dadaset & Mass & Axis-aligned & Annotated \\
\hline
SemanticKITTI & 61.6 & 61.7 & - \\
nuScenes & 72.6 & 74.7 & 74.8 \\
\Xhline{0.8pt}
\end{tabular}
\begin{center}
\vspace{-1.5em}
\caption{PQ results with different choices of instance center labels on SemanticKITTI and nuScenes validation.}
\label{tabcenter}
\end{center}
\vspace{-3.0em}
\end{center}
\end{table}

\section{Qualitative Results}

\begin{figure*}[t]
\begin{center}
\includegraphics[width=0.94\linewidth]{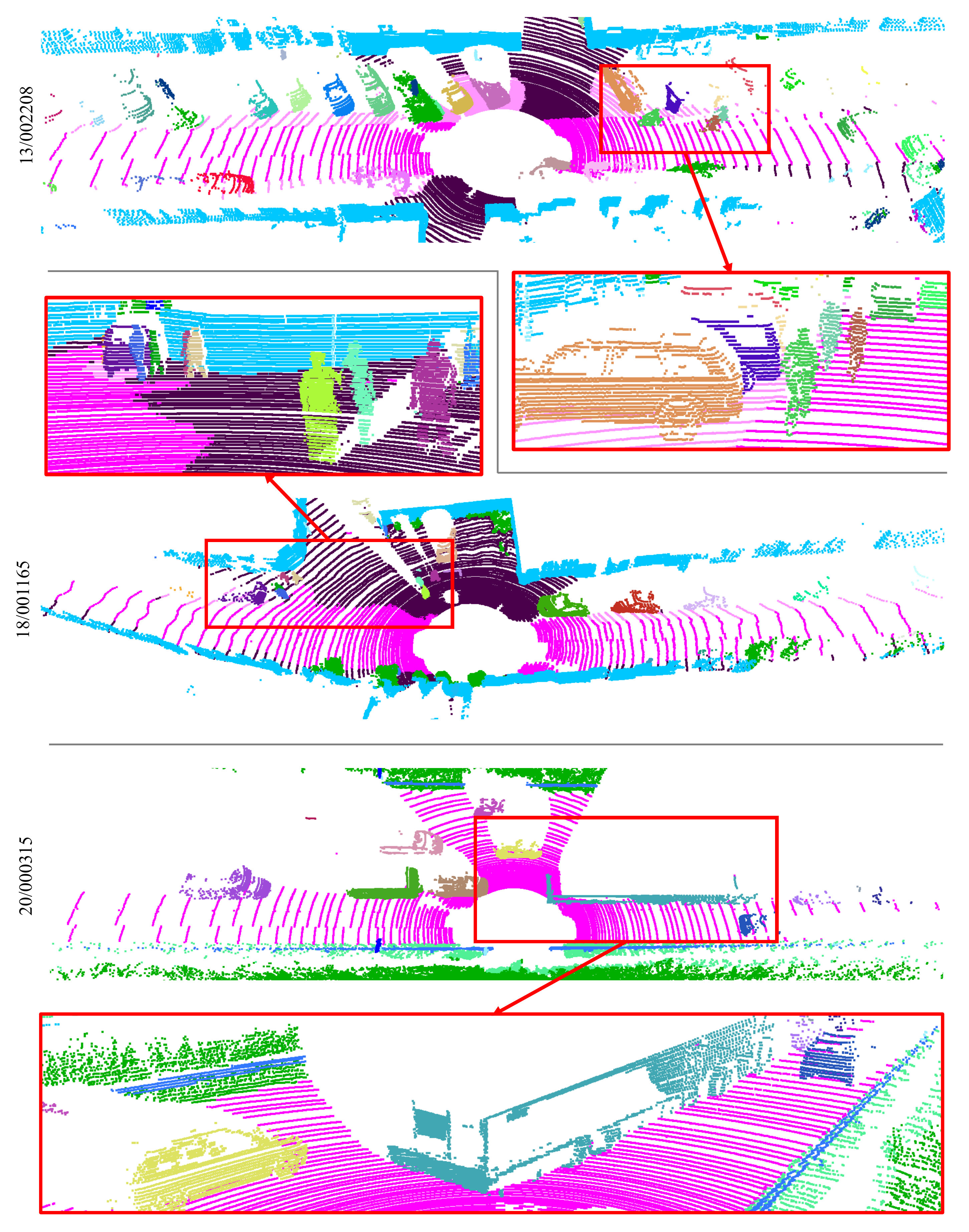}
\end{center}
   \caption{Qualitative examples on SemanticKITTI. The top two examples show the performance of our method in crowded scenes. The bottom example focuses on the big object segmentation.}
\label{fig_sub2}
\vspace{-0.5em}
\end{figure*}

\begin{figure*}[t]
\begin{center}
\includegraphics[width=\linewidth]{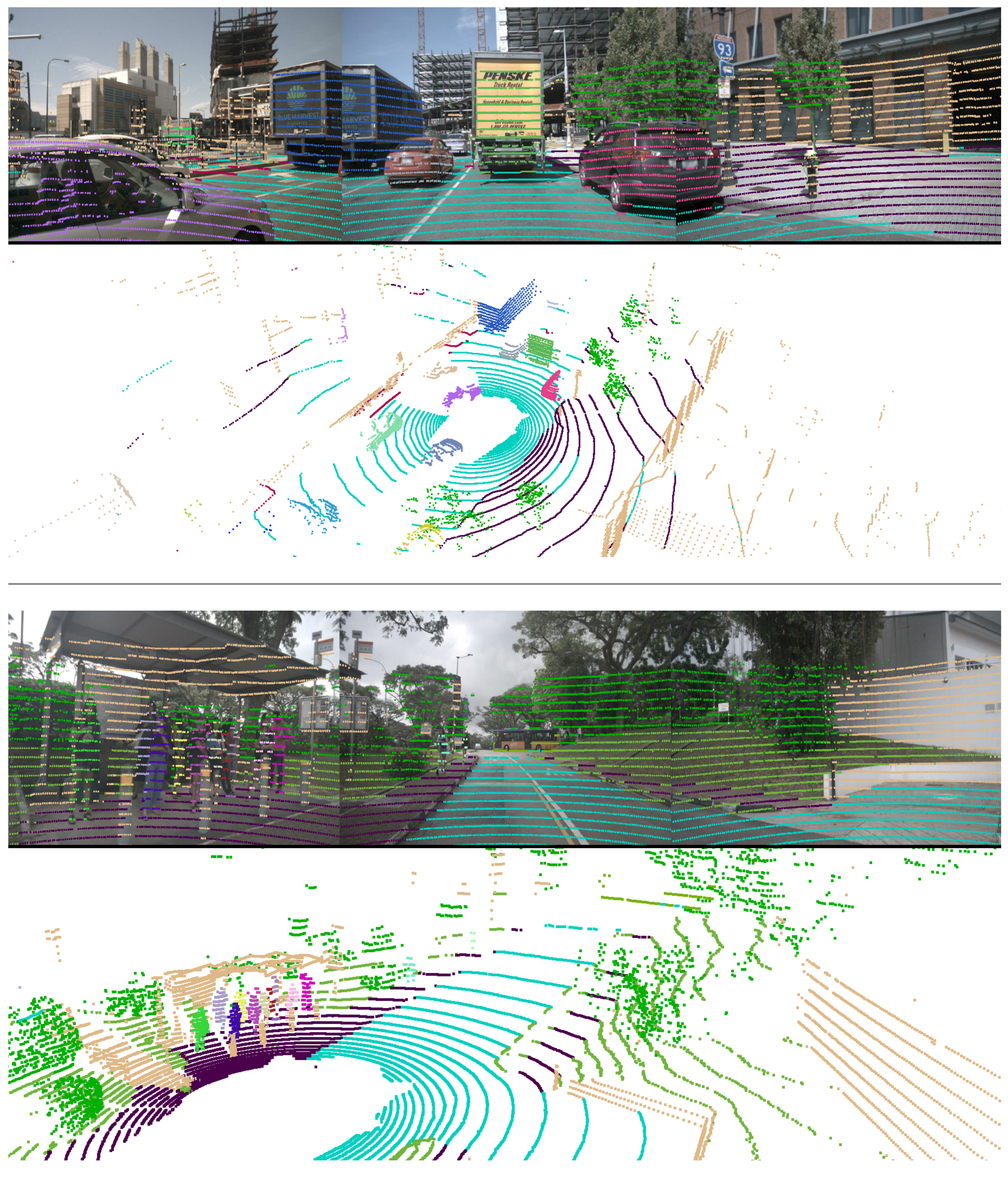}
\vspace{-2.5em}
\end{center}
   \caption{Qualitative examples on nuScenes. The two examples show a driving scene and a group of crowded people respectively.}
\label{fig_sub3}
\vspace{-0.5em}
\end{figure*}

\begin{figure*}[t]
\begin{center}
\includegraphics[width=\linewidth]{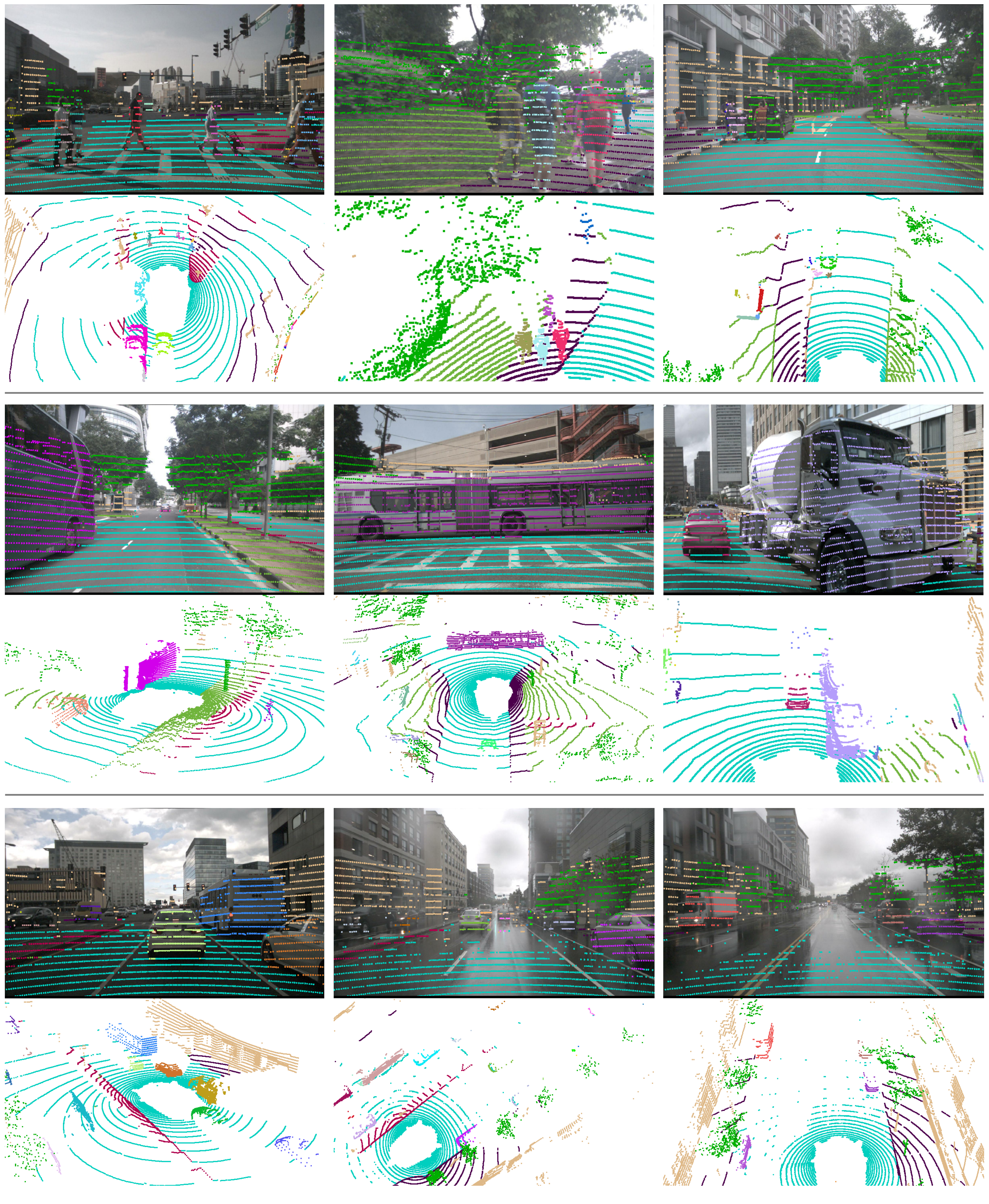}
\vspace{-2.5em}
\end{center}
   \caption{Qualitative examples on nuScenes, including small objects that are close to each other (row1), big objects (row2), as well as cloudy and rainy day (row3).}
\label{fig_sub4}
\vspace{-0.5em}
\end{figure*}

We show the visualization examples of our Panoptic-PHNet on SemanticKITTI in \cref{fig_sub2}, as well as on nuScenes in \cref{fig_sub3} and \cref{fig_sub4}. We use the official color map for \emph{stuff} regions and random colors for instance IDs. For nuScenes, we also project panoptic segmentation results onto the front view images. It can be observed that our approach performs well not only for crowded scenes, but also for big objects, which are the focuses of our paper while often ignored in previous studies. Specifically, as shown in the bottom image of \cref{fig_sub3}, a group of close persons are correctly segmented thanks to our high-quality offset regression and efficient clustering pseudo heatmap.


\section{Performance across Classes}

We show the detailed class-wise results of our Panoptic-PHNet on SemanticKITTI and nuScenes in \cref{tab_sup1}, \cref{tab_sup2} and \cref{tab_sup3}.

\newpage
\vspace{20em}

\newpage


\newpage
\newpage
\newpage

\begin{table*}[t]
\small
\begin{center}
\setlength{\tabcolsep}{1.2mm}
\begin{tabular}{c|ccccccccccccccccccc|c}
\hline

Metrics & \rotatebox{90}{Car} & \rotatebox{90}{Truck} & \rotatebox{90}{Bicycle} & \rotatebox{90}{Motorcycle} & \rotatebox{90}{Other Vehicle} & \rotatebox{90}{Person} & \rotatebox{90}{Bicyclist} & \rotatebox{90}{Motorcyclist} & \rotatebox{90}{Road} & \rotatebox{90}{Sidewalk} & \rotatebox{90}{Parking} & \rotatebox{90}{Other Ground} & \rotatebox{90}{Building} & \rotatebox{90}{Vegetation} & \rotatebox{90}{Trunk} & \rotatebox{90}{Terrain} & \rotatebox{90}{Fence} & \rotatebox{90}{Pole} & \rotatebox{90}{Traffic Sign} & Mean \\

\hline
PQ & 94.0 & 45.1 & 54.6 & 62.4 & 51.2 & 74.4 & 76.3 & 52.0 & 89.9 & 70.6 & 49.4 & 11.7 & 87.8 & 79.4 & 57.2 & 45.0 & 52.6 & 54.5 & 61.2 & 61.5 \\

RQ & 98.6 & 47.5 & 71.8 & 69.9 & 54.9 & 82.8 & 83.2 & 54.6 & 96.0 & 85.3 & 63.2 & 15.6 & 93.4 & 95.0 & 77.0 & 59.0 & 68.6 & 72.5 & 80.2 & 72.1 \\

SQ & 95.4 & 95.0 & 76.0 & 89.3 & 93.3 & 89.8 & 91.7 & 95.2 & 93.6 & 82.8 & 78.1 & 75.0 & 94.1 & 83.6 & 74.3 & 76.3 & 76.6 & 75.2 & 76.3 & 84.8 \\

IoU & 96.3 & 56.4 & 59.4 & 55.5 & 48.0 & 66.2 & 70.0 & 22.9 & 92.1 & 77.5 & 67.9 & 33.0 & 92.8 & 84.9 & 69.3 & 69.8 & 68.5 & 61.2 & 62.2 & 66.0\\
\hline
\end{tabular}
\end{center}
\vspace{-2.0em}
\caption{Class-wise LiDAR panoptic segmentation resuts on the \textbf{test} set of SemanticKITTI. All scores are in [$\%$].}
\label{tab_sup1}
\end{table*}

\begin{table*}[t]
\begin{center}
\setlength{\tabcolsep}{1.5mm}
\begin{tabular}{c|cccccccccccccccc|c}
\hline

Metrics & \rotatebox{90}{Barrier} & \rotatebox{90}{Bicycle} & \rotatebox{90}{Bus} & \rotatebox{90}{Car} & \rotatebox{90}{Construction Vehicle} & \rotatebox{90}{Motorcycle} & \rotatebox{90}{Pedestrian} & \rotatebox{90}{Traffic Cone} & \rotatebox{90}{Trailer} & \rotatebox{90}{Truck} & \rotatebox{90}{Driveable Surface} & \rotatebox{90}{Other Flat} & \rotatebox{90}{Sidewalk} & \rotatebox{90}{Terrain} & \rotatebox{90}{Manmade} & \rotatebox{90}{Vegetation} & Mean \\

\hline
PQ & 68.0 & 77.6 & 75.4 & 95.5 & 75.9 & 91.1 & 94.9 & 94.8 & 71.7 & 76.4 & 97.7 & 52.0 & 75.2 & 59.4 & 88.8 & 86.5 & 80.1\\
RQ & 82.4 & 86.4 & 78.9 & 97.9 & 82.5 & 96.1 & 99.0 & 99.1 & 78.8 & 80.2 & 100.0 & 59.2 & 90.0 & 75.5 & 98.6 & 96.5 & 87.6\\
SQ & 82.5 & 89.8 & 95.7 & 97.5 & 92.1 & 94.7 & 95.8 & 95.7 & 90.9 & 95.2 & 97.7 & 87.8 & 83.6 & 78.7 & 90.1 & 89.7 & 91.1\\
IoU & 84.3 & 35.6 & 84.9 & 93.1 & 70.1 & 88.0 & 82.0 & 81.1 & 86.6 & 73.2 & 97.7 & 68.4 & 80.6 & 76.1 & 92.2 & 88.7 & 80.2\\
\hline
\end{tabular}
\end{center}
\vspace{-2.0em}
\caption{Class-wise LiDAR panoptic segmentation resuts on the \textbf{test} set of nuScenes. All scores are in [$\%$].}
\label{tab_sup2}
\end{table*}

\begin{table*}[t]
\begin{center}
\setlength{\tabcolsep}{1.5mm}
\begin{tabular}{c|cccccccccccccccc|c}
\hline

Metrics & \rotatebox{90}{Barrier} & \rotatebox{90}{Bicycle} & \rotatebox{90}{Bus} & \rotatebox{90}{Car} & \rotatebox{90}{Construction Vehicle} & \rotatebox{90}{Motorcycle} & \rotatebox{90}{Pedestrian} & \rotatebox{90}{Traffic Cone} & \rotatebox{90}{Trailer} & \rotatebox{90}{Truck} & \rotatebox{90}{Driveable Surface} & \rotatebox{90}{Other Flat} & \rotatebox{90}{Sidewalk} & \rotatebox{90}{Terrain} & \rotatebox{90}{Manmade} & \rotatebox{90}{Vegetation} & Mean \\

\hline
PQ & 53.5 & 77.5 & 75.4 & 90.8 & 48.6 & 87.3 & 91.0 & 87.0 & 56.5 & 72.6 & 96.7 & 58.3 & 72.4 & 54.9 & 88.7 & 84.8 & 74.7\\
RQ & 67.7 & 89.4 & 80.6 & 95.5 & 60.5 & 95.0 & 97.4 & 95.2 & 65.4 & 78.6 & 99.8 & 67.8 & 88.0 & 69.6 & 99.0 & 97.0 & 84.2\\
SQ & 79.1 & 86.7 & 93.5 & 95.0 & 80.4 & 91.9 & 93.5 & 91.3 & 86.4 & 92.3 & 96.8 & 85.9 & 82.3 & 78.9 & 89.6 & 87.4 & 88.2\\
IoU & 77.9 & 52.4 & 93.5 & 93.0 & 57.0 & 88.1 & 83.9 & 69.9 & 69.6 & 86.3 & 96.9 & 75.3 & 76.3 & 75.3 & 90.7 & 88.7 & 79.7\\
\hline
\end{tabular}
\end{center}
\vspace{-2.0em}
\caption{Class-wise LiDAR panoptic segmentation resuts on nuScenes validation. All scores are in [$\%$].}
\label{tab_sup3}
\end{table*}

\end{document}